\documentclass[conference]{IEEEtran}
\IEEEoverridecommandlockouts

\usepackage{multirow}
\usepackage[ruled,vlined]{algorithm2e}
\usepackage{graphicx}
\usepackage[table]{xcolor} 
\usepackage{booktabs}

 \author{
 	\parbox{\textwidth}{%
 		\centering
 		Toshiya Kitahara, Ryu Shirakami, Koh Takeuchi, Hisashi Kashima
 	}%
 	\thanks{Toshiya Kitahara is with Sumitomo Electric System Solutions Co., Ltd.
    Ryu Shirakami, Koh Takeuchi and Hisashi Kashima are with Kyoto University.
    Corresponding author: Toshiya Kitahara (kitahara-toshiya@seiss.co.jp).
    }%
 }

\usepackage{amsmath,amssymb,amsfonts}
\usepackage{bm}
\usepackage{mathrsfs}
\usepackage{booktabs}
\usepackage{xcolor}
\usepackage{comment}
\usepackage[hidelinks]{hyperref}

\DeclareMathOperator*{\argmin}{argmin}
\newcommand{\bmx}[0]{\bm{X}} 
\newcommand{\bmX}[0]{\mathcal{X}}

\newcommand{\KT}[1]{\textcolor{black}{#1}}
\newcommand{\RS}[1]{\textcolor{black}{#1}}

\def\ul#1{\underline{#1}}

\title{\vspace{3mm}\LARGE
CASTANET:
Causality-Aware Spatio-Temporal Adversarial Network Using Traffic Incident Effects}

\begin{document}
\maketitle
\thispagestyle{empty}
\pagestyle{empty}

\begin{abstract}
Predicting non-periodic traffic congestion caused by sudden incidents (e.g., accidents and road damage) is crucial for advanced intelligent transportation systems.
However, incident-driven congestion is difficult to forecast because incidents are extremely sparse, occur at specific times and locations, and have heterogeneous impacts depending on the traffic context. While recent deep learning approaches have significantly improved periodic traffic forecasting, their performance on non-periodic congestion remains limited, partly because incident records are not explicitly incorporated and their occurrence is strongly biased in space and time. To address these challenges, we propose CASTANET, which integrates spatio-temporal graph neural networks and causal treatment effect estimation to utilize incident records while mitigating selection bias. Experiments on real-world traffic data and accident records from Tokyo, which we treat as incidents, show that CASTANET reduces RMSE by 4.0\% overall compared to the best baseline and by 10.1\% on incident-conditioned evaluation, with gains reaching 14.55\% under severe congestion.
\end{abstract}

\section{Introduction}
Traffic congestion management is an essential function of advanced intelligent transportation systems (ITS).
In urban road networks, congestion can be broadly categorized into periodic and non-periodic congestion~\cite{sakakibara1999moderato}.
Periodic congestion arises from regular social activities such as commuting and logistics, and therefore repeatedly occurs at similar times and locations.
In contrast, non-periodic congestion occurs irregularly and may propagate rapidly through the road network because sudden incidents, such as traffic accidents or road damage, often trigger such congestion. 
If the growth of such incident-driven congestion can be forecast, traffic operators can respond proactively, for example by adapting signal timings and allocating operational resources.
We also need to know whether the growth is mainly attributable to the incident or would have occurred even without it.

Although the causes of non-periodic congestion are varied, we hypothesize that incident-driven congestion growth is predictable by modeling the interaction between traffic conditions, road structure, and incidents.
Recently, spatio-temporal graph neural networks (STGNNs)~\cite{wu2019gwnt,li2018dcrnn,bai2020agcrn} have shown their potential to capture such complex relationships from large-scale traffic observations.
However, predicting non-periodic congestion remains challenging for two key reasons.

\begin{figure}[tb]
    \centering
    \includegraphics[width=1.0\linewidth]{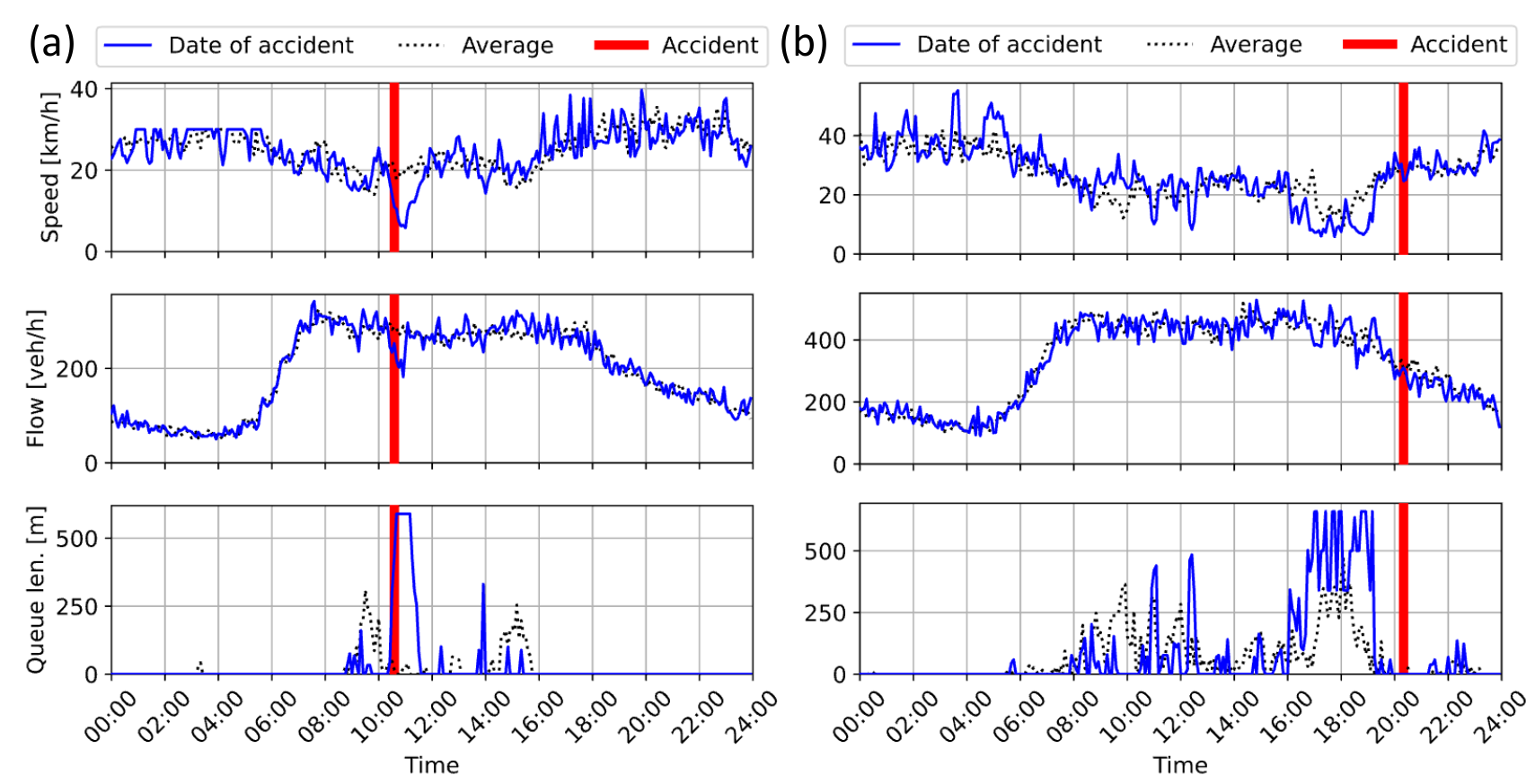}
\caption{Speed, flow, and queue length under two incidents. (a) Speed and flow drop while queue length rises after the incident. (b) No clear traffic changes are observed.}
\label{fig:data_accidents}
\end{figure}

First, the impact of a sudden incident is heterogeneous~\cite{LIN2020accidentimpact}.
We show traffic observations around incidents on two different roads in Figure~\ref{fig:data_accidents}.
In the left figure, an incident at around 10{:}30 causes a sharp decline in speed and flow, leading to a rapid increase in queue length.
In contrast, in the right figure, there are no apparent changes in traffic after the incident.
These contrasting outcomes arise because the impact of an incident depends on the incident itself, traffic conditions, and the local road network structure.
A predictive model must therefore capture these context-dependent interactions rather than treating all incidents uniformly.

\begin{figure}[t]
    \centering
    \includegraphics[width=0.9\linewidth]{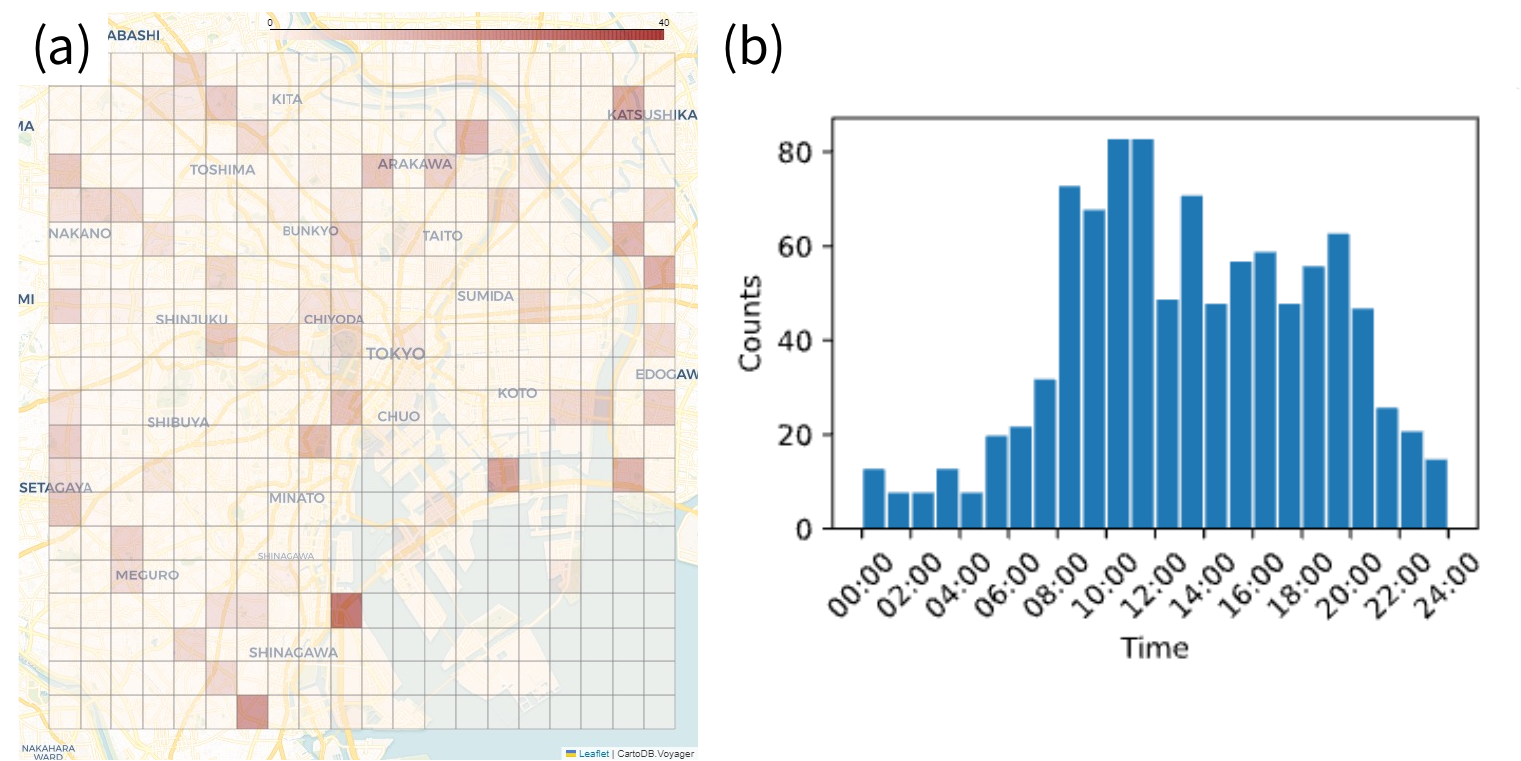}
    \caption{
    (a) The spatial distribution of incidents in Tokyo from May 1 to September 26,
    where darker meshes indicate higher frequencies. (b) Temporal histogram of incidents.
    }
    \label{fig:accidents}
\end{figure}

Second, incident data are inherently sparse and exhibit strong spatio-temporal biases~\cite{moosavi2019accidentrisk}. In our traffic dataset from Tokyo, incidents occurred in only $0.01\%$ of all samples. 
The fact that the locations and times where incidents occur are not uniformly distributed is an additional complexity.
Figure~\ref{fig:accidents} shows spatial and temporal histograms of incidents in Tokyo, indicating non-uniform distribution across time or space.
This sparsity makes prediction difficult, but a more subtle problem arises: incidents occur more frequently under specific conditions.
Naively adding incident indicators as input features to an STGNN therefore risks confounding the causal effect of an incident with the traffic conditions that make incidents more likely.
A model may attribute congestion to an incident even when similar congestion would have occurred without it, which is undesirable for ITS decision-making.
 
To mitigate this confounding, we draw on the treatment effect estimation framework from causal inference~\cite{rubin2005potentialoutcome}, where an incident can be viewed as a non-random intervention and the impact on congestion is the treatment effect.
A key technique developed in this framework is balanced representation learning~\cite{shalit2017tarnet}, which learns latent representations where treated and control groups are similarly distributed, thereby reducing selection bias and improving predictive accuracy even for factual outcomes.
These methods have been extended to multivariate time series~\cite{bica2020crn,CausalTransformer2022} and spatio-temporal data~\cite{jing-www22,takeuchi2023sinet}.
However, these methods have been developed for domains such as epidemiology~\cite{jing-www22}, and hierarchical spatial systems~\cite{takeuchi2023sinet}, and have not been applied to traffic congestion prediction, where incidents are extremely sparse, exhibit strong spatio-temporal selection bias, and interact with directional traffic propagation on road networks.

In this paper, we propose a causality-aware spatio-temporal adversarial network (CASTANET) for Incident-Aware traffic congestion prediction.
CASTANET predicts future congestion while accounting for traffic context, including traffic history, road structure, and records of sudden traffic incidents.
The key idea is to integrate STGNNs with treatment effect estimation, so that sudden incidents can be treated as non-random interventions that affect future traffic.
First, we combine an STGNN with balanced representation learning methods~\cite{bica2020crn} to learn latent representations of traffic conditions from historical traffic observations, incident history, and road structure~\cite{wu2019gwnt}.
Second, we predict future congestion using the learned representations with the current presence or absence of a sudden incident.
To mitigate bias due to spatio-temporally imbalanced incident occurrences, we incorporate adversarial representation learning (ARL)~\cite{bica2020crn,CausalTransformer2022}, which encourages balanced representations and reduces dependence on factors that merely make incidents more likely to occur.

We conduct experiments on a real-world dataset consisting of traffic measurements and incident records for approximately two thousand road segments in Tokyo, Japan.
We focus on predicting queue length, which is an important quantity for traffic signal control.
Experimental results show that CASTANET reduces queue length prediction RMSE by 4.0\% overall compared to the best external baseline (QTNet) and by 10.1\% in post-incident RMSE compared to the best ablation variant, with gains reaching 14.55\% in severe congestion against state-of-the-art models.

We summarize the contributions of this study as follows:
\begin{itemize}
  \item \textbf{Causality-aware formulation of incident-driven congestion prediction:}
  We formulate Incident-Aware traffic prediction within the treatment effect estimation framework, leveraging balanced representation learning to reduce the influence of confounding factors correlated with both incidents and congestion.

  \item \textbf{Integration of STGNNs with adversarial balanced representations:}
  We develop CASTANET, which combines an STGNN encoder with adversarial balanced representation learning tailored to the challenges of traffic incident data, including extreme sparsity and strong spatio-temporal selection bias in incident occurrences.

  \item \textbf{Empirical validation on large-scale real-world data:}
  Using seven months of traffic and incident records in Tokyo, we demonstrate that CASTANET consistently outperforms state-of-the-art baselines, with particularly large improvements in post-incident and severe-congestion scenarios where accurate prediction matters most for operational decisions.
\end{itemize}

\section{Related Work}
\subsection{Traffic Forecasting}
Traffic engineering has developed mathematical models to simulate traffic flow by assuming theoretical conditions.
Macroscopic models, such as queueing theory~\cite{cascetta2013transportation} and kinematic wave theory~\cite{daganzo2003kw}, attempt to simulate large-scale traffic.
Microscopic models, such as car-following models~\cite{olstam2004sims}, aim to simulate individual vehicles' behavior.
However, real-world traffic data often violate the theoretical assumptions, and do not accurately match the simulated results. Recently, data-driven methods have become popular in traffic prediction~\cite{yin2021survey}.
Researchers have attempted to capture spatio-temporal dependencies by combining GNNs
for the spatial domain, and a recurrent neural network (RNN) for the temporal domain.
The models have demonstrated improvements on the traffic prediction~\cite{li2018dcrnn,wu2019gwnt,bai2020agcrn,jiang2023megacrn}, but \KT{they do not incorporate} incident data.
Various methods for predicting future incidents from traffic data have been developed and have successfully improved their performance~\cite{moosavi2019accidentrisk} because incidents tend to occur under certain traffic conditions.
These results suggest that incidents are not random in space and time.
In this context, studies have considered the impact of incidents on traffic~\cite{yu2017lstm,LIN2020accidentimpact}. 
However, using traffic incident data to predict future traffic is out of their focus. 
\subsection{Causal Inference}
Treatment effect estimation in causal inference is a method for estimating the change in outcomes due to the selection of treatment \cite{rubin1985}.
For example, this corresponds to estimating how much the symptom of a disease changes when medication is given to a patient or not. This framework introduces the notion of missing counterfactual outcomes and accounts for selection bias when treatments are not assigned at random~\cite{rubin2005potentialoutcome}.
Researchers have proposed bias-reduction methods to accurately estimate average treatment effects given a condition~\cite{curth2021nonparametric}.
Econometrics and statistics have extended this framework for spatial data analysis \cite{akbari2024spatial}. Recently, deep learning methods for estimating CATE (Conditional Average Treatment Effect) have emerged and received considerable attention in machine learning~\cite{bica2021survey}.
Their core idea is to remove the selection bias by learning representations where treatment and control groups are identically distributed~\cite{johansson2016ipm,shalit2017tarnet}.
Since removing bias can prevent overfitting to factual outcomes, they have demonstrated the improved accuracy for not only CATE but for predicting factual outcomes.
Various methods have been developed for estimating CATE on multivariate series ~\cite{bica2020crn,CausalTransformer2022},  and spatio-temporal data~\cite{jing-www22,takeuchi2023sinet}.
For example, they attempted to consider the spread of \KT{diseases}, the behavior of agents, and the evacuation movements of crowds, but traffic data is out of their scope.
Existing studies have utilized causal relationships between time-series data and a road network for graph structure learning~\cite{zhang2022grangertraffic,lin2023dynamic} or introduce causal inference to capture relationships between multi-modal time-series data~\cite{Zhao2023causaltraffic}. 
However, these models have not considered sudden traffic incidents and their effect on traffic.
\section{Problem Setting}\label{sec:problemsetting}
We represent a road network by an adjacency matrix $\bm{W} \in \{0,1\}^{N \times N}$, where $N$ is the number of road segments.
The element $w_{ij}$ is set to 1 if segments $i$ and $j$ are connected, or 0 otherwise.
Let $\bmx^{\text{traffic}}_{t} \in \mathbb{R}^{N \times d_{\text{traffic}}}$ denote traffic observations at time step $t$, where $d_{\text{traffic}}$ is the number of traffic variables.
$\bmx^{\text{traffic}}_{t}$ includes queue length, speed, and flow for each segment.
We use auxiliary features $\bmx^{\text{aux}}_{t} \in \mathbb{R}^{N \times d_{\text{aux}}}$, such as time-of-day, day-of-week, and static road attributes.
We concatenate them as $\bmx_t = [\bmx^{\text{traffic}}_{t},\, \bmx^{\text{aux}}_{t}] \in \mathbb{R}^{N \times d_{\text{feature}}}$, where $d_{\text{feature}} = d_{\text{traffic}} + d_{\text{aux}}$.

We represent the current incident status by a binary vector $\bm{a}_t \in \{0,1\}^{N}$, where $a_{i,t}=1$ if an incident occurs on segment $i$ at time $t$, or $a_{i,t}=0$ otherwise.
We denote the traffic history over the last $T$ time steps by
$\bmX_t = [\bmx_{t-T+1}, \ldots, \bmx_t] \in \mathbb{R}^{N \times d_{\text{feature}} \times T}$,
and the incident history (excluding the current time step) by
$\bm{A}_{t-1} = [\bm{a}_{t-T}, \ldots, \bm{a}_{t-1}] \in \{0,1\}^{N \times T}$.
Let $\bm{y}_t \in \mathbb{R}^{N}$ denote the target congestion variable at time $t$ (queue length in our experiments), and define the prediction target for the next $T'$ time steps as
$\bm{Y}_t = [\bm{y}_{t+1}, \ldots, \bm{y}_{t+T'}] \in \mathbb{R}^{N \times T'}$.

Given a dataset $D=\{(\bmX_t, \bm{A}_{t-1}, \bm{a}_t, \bm{Y}_t,\, \bm{W})\}_{t=1}^{\tau}$ with $\tau$ time steps, our goal is to learn a predictor $f_{\theta}$ that outputs $\hat{\bm{Y}}_t$:
\begin{equation}
\hat{\bm{Y}}_t = f_{\theta}(\bmX_t, \bm{A}_{t-1}, \bm{a}_t,\, \bm{W}).
\label{eq:problem_pred}
\end{equation}
We estimate parameters $\hat{\theta}$ by minimizing the empirical risk with a loss function $\ell(\cdot,\cdot)$:
\begin{equation}
\hat{\theta} = \arg\min_{\theta}\frac{1}{\tau}\sum_{t=1}^{\tau}\ell(\bm{Y}_t,\hat{\bm{Y}}_t).
\label{eq:problem_erm}
\end{equation}

\section{CASTANET}
As shown in Fig.~\ref{fig:accidents}, sudden incidents such as traffic accidents are not uniformly distributed across time and space. To mitigate this issue, we build on the treatment effect estimation framework~\cite{robins2008,rubin2005potentialoutcome}.

\subsection{Causal Viewpoint}
\begin{figure}[t]
    \centering
    \includegraphics[width=0.4\linewidth]{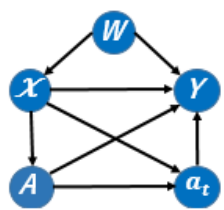}
    \caption{A causal graph representing the relationship between traffic data and incidents.
    $\bmX$, $\bm{A}$, $\bm{Y}$, $\bm{a}_t$, and $\bm{W}$ denote the traffic history, incident history, outcome, current incident status, and road network, respectively.}
    \label{fig:causal_graph}
\end{figure}

We treat the traffic history $\bmX_t$, incident history $\bm{A}_{t-1}$, and road network $\bm{W}$ as covariates, the current incident status $\bm{a}_{t}$ as a treatment, and the future traffic $\bm{Y}_t$ as the outcome.
We assume that these variables follow the causal graph shown in Fig.~\ref{fig:causal_graph}.
The outcome $\bm{Y}_t$ depends on $\bmX_t$, $\bm{A}_{t-1}$, $\bm{W}$, and $\bm{a}_{t}$, while the occurrence of incidents is conditioned on recent covariates.
For simplicity, we assume that incidents occur independently across road segments.
We define the propensity score for $a_{i,t}$ as the conditional probability:
\begin{align} \label{eq:propensity}
    P(a_{i,t}=1 \mid \bmX_t,\bm{A}_{t-1}) = s_{i,t}.
\end{align}
Following~\cite{bica2020crn,CausalTransformer2022}, we assume three basic conditions:

\textbf{Assumption 1: Consistency.}
If the realized treatment for segment $i$ at time $t$ is $a_{i,t}$, then the observed outcome equals the corresponding potential outcome under $a_{i,t}$.

\textbf{Assumption 2: Positivity.}
The propensity score satisfies $0 < P(a_{i,t}=1 \mid \bmX_t,\bm{A}_{t-1}) < 1$.

\textbf{Assumption 3: Sequential strong ignorability.}
Given the sequence of recent covariates, the current treatment assignment is conditionally independent of future potential outcomes. 

\begin{figure}
    \centering
    \includegraphics[width=0.9\linewidth]{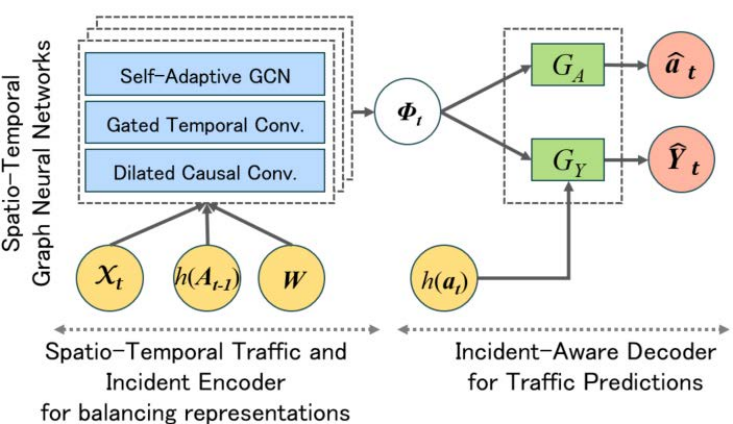}
    \caption{The overview of CASTANET. 
    Incident embedding $h$ transforms the incident history $\bm{A}_{t-1}$ and the current incident status $\bm{a}_{t}$.
    Spatio-temporal encoder uses STGNNs to extract latent representation $\bm{\Phi}_t$ from traffic data $\bmX_t$ and embedding $h(\bm{A}_{t-1})$ using an adjacency matrix $\bm{W}$.
    Decoder predicts future traffic $\hat{\bm{Y}_t}$ and the current incident status $\hat{\bm{a}_t}$ with $G_Y$ and $G_A$.
    }
    \label{fig:model}
\end{figure}

Based on this viewpoint, we propose a causality-aware spatio-temporal adversarial network (CASTANET) that incorporates incident information to predict future traffic.
CASTANET integrates spatio-temporal graph neural networks (STGNNs)~\cite{wu2019gwnt} and treatment effect estimation.
Figure~\ref{fig:model} shows an overview of CASTANET.
We extend deep representation learning methods for multivariate time series~\cite{bica2020crn,CausalTransformer2022} to incorporate spatial traffic relationships through the road network $\bm{W}$.

\subsection{Spatio-Temporal Traffic and Incident Encoder} \label{sec:embedding}
\textbf{Incident embedding:}
We represent the incident status of segment $i$ at time step $t$ by a binary indicator $a_{i,t}$, which may be too limited to capture the relationship between incidents and traffic.
We therefore embed the incident status into a learnable vector space~\cite{takeuchi2023sinet}:
\begin{align}
    h(a_{i,t}) =
    \begin{cases}
        \bm{u}^{(0)} & \text{if } a_{i,t}=0,\\
        \bm{u}^{(1)} & \text{if } a_{i,t}=1,
    \end{cases}
\end{align}
where $\bm{u}^{(0)},\bm{u}^{(1)}\in\mathbb{R}^{d_{\rm emb}}$ are learnable $d_{\rm emb}$-dimensional vectors.
We obtain incident embeddings from the historical records $\bm{A}_{t-1}$ as
$h(\bm{A}_{t-1})=[h(\bm{a}_{t-T}), \dots, h(\bm{a}_{t-1})]\in\mathbb{R}^{N \times d_{\rm emb} \times T}$,
and from the current incident status $\bm{a}_t$ as $h(\bm{a}_t) \in\mathbb{R}^{N \times d_{\rm emb}}$.

\textbf{Latent spatio-temporal representation:}\label{sec:repre}
We develop a spatio-temporal encoder to extract temporal and spatial contexts from historical traffic data, incident history, and the road network.
Following balancing representation methods~\cite{bica2020crn}, an STGNN $f_{\rm STGNN}$ extracts a latent representation $\bm{\Phi}_t\in\mathbb{R}^{N\times d_{\rm rep}}$ from $\bmX_t$, $h(\bm{A}_{t-1})$ and $\bm{W}$:
\begin{align}
  \bm{\Phi}_t = f_{\rm STGNN}(\bmX_t, h(\bm{A}_{t-1}), \bm{W}; \theta_{R}),
\end{align}
where $\theta_{\rm R}$ is a set of STGNN parameters.
The representation $\bm{\Phi}_t$ is designed to be predictive of the future outcome $\bm{Y}_t$ while being less predictive of the current incident status $\bm{a}_t$, which helps mitigate bias in incident occurrences.

We employ Graph WaveNet (GWNet)~\cite{wu2019gwnt} as the STGNN backbone, as it learns a self-adaptive adjacency matrix and captures dependencies beyond the predefined graph structure, which is particularly beneficial for complex road networks such as Tokyo. 
We concatenate the traffic history $\bmX_t$ and incident embeddings $h(\bm{A}_{t-1})$ as $[\bmX_t, h(\bm{A}_{t-1})]$ and use it as input to GWNet.
GWNet consists of $K$ recursive temporal and spatial layers.
The temporal layer applies a dilated causal convolution to extract historical features using only data prior to a given time, and a gated temporal convolution to adaptively select features.

The spatial layer employs graph convolution networks (GCNs)~\cite{kipf2017gcn} with both predefined and learned adjacency matrices.
Let $\bm{E}_1,\bm{E}_2\in\mathbb{R}^{N\times d_{\rm node}}$ be learnable node embeddings.
The learned adjacency matrix is computed as
$\bm{P}=\text{Softmax}(\text{ReLU}(\bm{E}_1\bm{E}_2^\top))$.
By learning $\bm{P}$, we mitigate the mismatch between real traffic propagation patterns and the hand-crafted graph $\bm{W}$.
We denote the self-adaptive GCN as
\begin{align} \label{eq:gcn}
    f_{\rm SAGCN}(\bm{H}) = \sum_{k=0}^K \bigl(\bm P^k_{\rm f}\,\bm{H}\,\bm{S}_{k1} + \bm P^k_{\rm b}\,\bm{H}\,\bm{S}_{k2} + \bm{P}^k\,\bm{H}\,\bm{S}_{k3}\bigr),
\end{align}
where $\bm{H}\in\mathbb{R}^{N\times d_{\rm rep}}$ is an input feature matrix, $\bm{S}_{k1}$, $\bm{S}_{k2}$, $\bm{S}_{k3} \in\mathbb{R}^{d_{\rm rep}\times d_{\rm rep}}$ are learnable parameters and $\bm{P}_{\rm f}, \bm{P}_{\rm b}\in\mathbb{R}^{N\times N}$ are row/column-normalized matrices derived from $\bm{W}$.

\subsection{Incident-Aware Decoder}
Our decoder uses the representation $\bm{\Phi}_t$ and the current incident embedding $h(\bm{a}_t)$ to predict future traffic $\hat{\bm{Y}}_t$.
We build an outcome predictor network $G_{Y}$ using a self-adaptive GCN and multilayer perceptrons (MLPs).
We first apply Eq.~\eqref{eq:gcn} to propagate the current incident information to neighbors:
\begin{align}\label{eq:GCN2}
    \bm{\Psi}_t = f_{\rm SAGCN}([\bm{\Phi}_t, h(\bm{a}_t)]).
\end{align}

When the target $\hat{\bm{Y}}_t$ is queue length, we employ a traffic engineering-informed layer called the queueing theory (QT) layer~\cite{shirakami2023qtnet} and MLPs within $G_Y$:
\begin{align}\label{eq:QT}
    \hat{\bm{V}}_t, \hat{\bm{Q}}_t, \hat{\bm{C}}_t = f_{\rm MLP1}([\bm{\Phi}_t, \bm{\Psi}_t]), \quad
    \hat{\bm{Y}}_t = f_{\rm QT}(\hat{\bm{V}}_t, \hat{\bm{Q}}_t, \hat{\bm{C}}_t),
\end{align}
where $\hat{\bm{V}}_t \in \mathbb{R}^{N \times T'}$, $\hat{\bm{Q}}_t \in \mathbb{R}^{N \times T'}$, and $\hat{\bm{C}}_t \in \mathbb{R}^{N \times T'}$ are the predicted speed, flow, and correction terms.
We denote by $\theta_{\rm Y}$ the parameters in Eq.~\eqref{eq:GCN2} and Eq.~\eqref{eq:QT}.

In addition, we build a treatment classification network $G_A$ to predict the current incident status from $\bm{\Phi}_t$ by employing MLPs:
\begin{align}
    \hat{\bm{a}}_t=\sigma\bigl(f_{\rm MLP2}(\bm{\Phi}_t; \theta_{A})\bigr),
\end{align}
where $\sigma(\cdot)$ is the sigmoid function and $\theta_{A}$ is learnable parameters.

\subsection{Adversarial Representation Learning (ARL)} \label{sec:arl}
To mitigate the spatio-temporal selection bias and extreme sparsity of incident records, we incorporate an adversarial objective that discourages the latent representation $\bm{\Phi}_t$ from encoding the current incident status $\bm{a}_t$.
Following~\cite{CausalTransformer2022}, we implement this idea using the counterfactual domain confusion (CDC) loss.
This helps (a) improve prediction accuracy for future traffic and (b) reduce reliance on factors that merely make incidents more (or less) likely under specific conditions.

For objective (a), we introduce a loss function $\mathcal{L}_Y(\theta_Y, \theta_R)$ for the outcome predictor $G_Y$.
We use queue length as the main target variable and speed/flow as auxiliary outputs.
Following~\cite{shirakami2023qtnet}, we define the per-time-step losses as
\begin{align} \label{rq:loss_detail_1}
    &\mathcal{L}_{t,\rm q} = \sum_{i=1}^N\sum_{k=1}^{T'} \frac{r_{i,k}}{NT'}\bigl|Y ^{t}_{i,k} - \hat{Y ^{t}}_{i,k}\bigr|,
\end{align}
\begin{align} \label{rq:loss_detail_2}
    \mathcal{L}_{t,\rm nq} = \sum_{i=1}^N\sum_{k=1}^{T'} \frac{r_{i,k}}{NT'}\frac{\bigl|Y ^{t}_{i,k} - \hat{Y ^{t}}_{i,k}\bigr|}{L_{i}},
\end{align}
\begin{align} \label{rq:loss_detail_3}
    &\mathcal{L}_{t,\rm s} = \sum_{i=1}^N\sum_{k=1}^{T'} \frac{1}{NT'}\frac{\bigl|V ^{t}_{i,k} - \hat{V ^{t}}_{i,k}\bigr|}{V ^{t}_{i,k}},
\end{align}
\begin{align} \label{rq:loss_detail_4}
    \mathcal{L}_{t,\rm f} = \sum_{i=1}^N\sum_{k=1}^{T'} \frac{1}{NT'}\bigl|Q ^{t}_{i,k} - \hat{Q ^{t}}_{i,k}\bigr|,
\end{align}
where $\mathcal{L}_{t,\rm q}$, $\mathcal{L}_{t,\rm nq}$, $\mathcal{L}_{t,\rm s}$, and $\mathcal{L}_{t,\rm f}$ are the losses for queue length, normalized queue length (segment congestion rate), speed, and flow, respectively.
$L_{i}$ denotes the length of the $i$-th road segment.
$Y ^{t}$, $V ^{t}$ and $Q ^{t}$ are the ground truth values for
queue length, speed and flow respectively.
We define the weight $r_{i,k}$~\cite{shirakami2023qtnet} based on queue length as:
$r_{i,k} = r_{0}$ if $Y ^{t}_{i,k}=0$, and
$r_{i,k} = \sigma \!\left(\frac{Y ^{t}_{i,k}-\delta_{1}}{\delta_{2}}\right)$ otherwise,
where $r_{0}$, $\delta_{1}$, and $\delta_{2}$ are hyperparameters.
A smaller $\delta_{1}$ assigns higher weights to moderate queues, whereas a larger $\delta_{1}$ shifts the emphasis toward severe/long queues; $\delta_{2}$ controls the sharpness of this transition. In addition, $r_{0}$ is the weight for zero-queue samples, introduced to account for the zero-inflated nature of queue-length data (as also noted in~\cite{shirakami2023qtnet}) and to prevent the abundant zero cases from dominating training.

We then define $\mathcal{L}_Y(\theta_Y, \theta_R)$ as a weighted sum:
\begin{align} \label{eq:loss_y}
    \mathcal{L}_{Y}(\theta_Y, \theta_R)
    = \sum_{t=1}^{\tau}\Bigl(
    \lambda_{1}\mathcal{L}_{t, \rm q}
    + \lambda_{2}\mathcal{L}_{t, \rm nq}
    + \lambda_{3}\mathcal{L}_{t, \rm s}
    + \lambda_{4}\mathcal{L}_{t, \rm f}
    \Bigr),
\end{align}
where $\lambda_i (i=1,\dots,4)$ are hyperparameters for scale adjustment.

To reduce sensitivity to sparse and biased incidents, we employ a counterfactual domain confusion loss~\cite{CausalTransformer2022}.
This objective trains the treatment classification network $G_A$ while encouraging the representation $\bm{\Phi}_t$ to be less predictive of the current treatment.
We use a weighted cross-entropy loss~\cite{he2009imbalanced}:
\begin{align} \label{eq:loss_a}
     \mathcal{L}_{A}(\theta_A, \theta_R) = & -\frac{1}{\tau N}\sum_{t=1}^{\tau}\sum_{i=1}^N\Bigl[\alpha_1a_{i,t}\log \hat{a}_{i,t} \nonumber \\ 
     & \quad + \alpha_2(1-a_{i,t})\log (1-\hat{a}_{i,t})\Bigr],
\end{align}
where $\alpha_1$ and $\alpha_2$ are hyperparameters to handle the rarity of incidents.

We further encourage \emph{unpredictability} by measuring the cross-entropy between outputs of the treatment classification network and a uniform distribution:

\begin{align} \label{eq:loss_conf}
    \mathcal{L}_{A}'(\theta_A, \theta_R)
    = -\frac{1}{\tau N}\sum_{t=1}^{\tau}\sum_{i=1}^{N}
    \Bigl[\tfrac12\log\hat{a}_{i,t} + \tfrac12\log(1-\hat{a}_{i,t})\Bigr].
\end{align}

Based on Eqs.~\eqref{eq:loss_y}, \eqref{eq:loss_a}, and \eqref{eq:loss_conf}, we train the model by iteratively updating $\theta_Y$, $\theta_R$, and $\theta_A$~\cite{CausalTransformer2022}:
\begin{align}
    (\hat{\theta}_Y, \hat{\theta}_R)
    &= \argmin_{\theta_Y, \theta_R}
    \mathcal{L}_{Y}(\theta_Y, \theta_R)
    + \lambda_{\rm Conf}\,\mathcal{L}_{A}'(\hat{\theta}_A, \theta_R),\\
    \hat{\theta}_A
    &= \argmin_{\theta_A}
    \mathcal{L}_{A}(\theta_A, \hat{\theta}_R),
\end{align}
where $\lambda_{\rm Conf}$ is a hyperparameter controlling the strength of domain confusion.

\def\ul#1{\underline{#1}}

\section{Experiments}

We conducted experiments using real-world data to evaluate the queue length prediction performance of CASTANET. 
In the experiments, we aim to answer the following four research questions. 
\textbf{RQ1:} Does CASTANET outperform state-of-the-art baseline models in queue length prediction, particularly in post-incident scenarios?
\textbf{RQ2:} Does each component of CASTANET contribute to \KT{improving} the prediction \RS{performance}?
\textbf{RQ3:} How does the proposed model actually predict traffic congestion?
\textbf{RQ4:} How does the proposed model scale when applied to a significantly larger road network? 

\subsection{Dataset}
We used seven months of traffic and accident data collected by the Tokyo traffic control system from May 1, 2022, to November 30, 2022. 
In the dataset, only accidents were considered as incidents that affect future traffic. 
The targeted road network includes $N=2{,}032$ road segments in central Tokyo, covering a total of $1{,}468$ km. 
The traffic data \KT{include} three variables ($d_{\rm traffic}=3$: queue length, speed, and flow for each segment). 
Although the data \KT{were} collected at 50 s intervals, \KT{they} contain significant noise. Therefore, we downsampled \KT{them} to 5 min intervals and applied smoothing.

The data were processed by the traffic control system based on information from road sensors. The system estimates vehicle density over the road network from densities measured directly by the sensors and flags areas with densities exceeding a predefined threshold as congested. This allows the identification of congested areas within the entire road network, which is then used to calculate queue length. In traffic signal control, it is necessary to consider the number of cars in a queue at each intersection. However, sparsely installed sensors make the accurate estimation of this variable infeasible, \KT{so} signal control is based on queue lengths~\cite{sakakibara1999moderato,miyata1995stream}.

To evaluate robustness, we constructed five datasets using data from the earliest date spanning (a) 3, (b) 4, (c) 5, (d) 6, and (e) 7 months, and \KT{independently} trained and evaluated models for each. Each dataset was split in a 7:1:2 ratio in chronological order for training, validation, and test. Incident \RS{status} is denoted by ``1'' at the time and location of a registered incident; all other data points are denoted by ``0.'' 

The model input includes traffic data from the last $T=12$ timestamps (1 h), additional features, and current incident status. 
The output is the queue length for the next $T'=12$ timestamps (1 h). 
The additional features consist of the information about the periodicity and heterogeneity of the data, following~\cite{shirakami2023qtnet}. We utilized the average of the three traffic variables on each road segment. We also employed the information about each road segment, including the segment length, the number of inflow/outflow segments, the number of lanes and the number of intersections.

{\tabcolsep=0.2em
\begin{table*}[t]
   \centering
        \caption{
        RMSE and RMSE (Incident) for one-hour-ahead queue length predictions. The best-performing scores are shown in bold, and the second-best scores are \underline{underlined}. Parentheses indicate standard deviation. All models receive the same incident indicators as additional channels.
        }
    \label{tab:rmse_basleine}
    \scalebox{0.95}{
    \begin{tabular}{c c >{\columncolor{gray!15}}c c >{\columncolor{gray!15}}c c >{\columncolor{gray!15}}c c >{\columncolor{gray!15}}c}
    \toprule
        Model & \multicolumn{2}{c}{All} & \multicolumn{2}{c}{Non-zero} & \multicolumn{2}{c}{Top 10\%} & \multicolumn{2}{c}{Top 5\%} \\
        \cmidrule(lr){2-3}\cmidrule(lr){4-5}\cmidrule(lr){6-7}\cmidrule(lr){8-9}
        & RMSE[m] & RMSE & RMSE[m] & RMSE & RMSE[m] & RMSE & RMSE[m] & RMSE \\
        &  & (Incident)[m] &  & (Incident)[m] &  & (Incident)[m] &  & (Incident)[m]\\
        \midrule
        HA & $47.80_{(2.18)}$ & $205.76_{(42.45)}$ & $90.23_{(3.38)}$ & $239.63_{(47.74)}$ & $243.49_{(10.05)}$ & $417.13_{(68.78)}$ & $308.90_{(13.02)}$ & $474.00_{(85.49)}$\\
        Lasso & $177.70_{(15.07)}$ & $247.09_{(46.16)}$ & $249.15_{(17.08)}$ & $257.86_{(43.86)}$ & $325.30_{(21.36)}$ & $387.19_{(62.64)}$ & $373.92_{(23.85)}$ & $436.77_{(80.62)}$\\
        Ridge & $171.62_{(15.43)}$ & $246.51_{(46.20)}$ & $243.45_{(20.12)}$ & $331.64_{(11.09)}$ & $317.42_{(18.54)}$ & $407.45_{(52.52)}$ & $364.43_{(17.24)}$ & $434.84_{(79.94)}$\\
        XGBoost & $64.69_{(6.04)}$ & $224.08_{(16.30)}$ & $118.66_{(9.24)}$ & $260.52_{(15.74)}$ & $266.66_{(12.26)}$ & $351.46_{(64.31)}$ & $328.29_{(12.37)}$ & $392.32_{(72.39)}$\\
        \midrule
        DCRNN & $45.25_{(2.82)}$ & $181.39_{(42.98)}$ & $88.27_{(3.53)}$ & $210.98_{(47.75)}$ & $236.81_{(5.52)}$ & $331.81_{(64.15)}$ & $295.22_{(6.84)}$ & $368.97_{(77.92)}$\\
        GWNet & $44.15_{(2.67)}$ & $180.06_{(17.79)}$ & $86.18_{(3.30)}$ & $209.93_{(19.11)}$ & $230.12_{(5.79)}$ & $316.01_{(31.32)}$ & $286.51_{(7.48)}$ & $353.23_{(39.34)}$\\
        AGCRN & $45.10_{(2.95)}$ & ${178.76}_{(40.74)}$ & $88.05_{(3.86)}$ & ${207.90}_{(44.19)}$ & $237.98_{(7.91)}$ & $333.11_{(64.76)}$ & $297.38_{(10.45)}$ & $365.52_{(79.39)}$\\
        MegaCRN & $44.99_{(2.85)}$ & $196.21_{(28.59)}$ & $87.75_{(3.55)}$ & $228.43_{(29.45)}$ & $235.60_{(4.75)}$ & $367.46_{(51.53)}$ & $294.13_{(5.59)}$ & $386.57_{(44.87)}$\\
        QTNet & $\ul{43.57}_{(2.45)}$ & $192.05_{(44.50)}$ & ${84.23}_{(2.84)}$ & $223.24_{(49.91)}$ & ${207.78}_{(3.05)}$ & $315.91_{(89.22)}$ & ${259.64}_{(4.07)}$ & $347.96_{(108.05)}$\\
        PDG2Seq & $46.30_{(1.27)}$ & $184.59_{(21.89)}$ & $90.58_{(3.43)}$ & $215.47_{(26.67)}$ & $241.29_{(12.48)}$ & $\ul{287.72}_{(65.21)}$ & $300.95_{(14.30)}$ & $\ul{316.04}_{(85.18)}$\\
        
        STAEFormer & $45.31_{(2.69)}$ & $182.11_{(25.26)}$ & $88.29_{(3.30)}$ & $212.28_{(27.52)}$ & $235.56_{(7.66)}$ & $316.02_{(71.40)}$ & $293.72_{(10.53)}$ & $342.91_{(91.27)}$\\
        \midrule
        CASTANET:NoARL1 & ${43.60}_{(2.48)}$ & $\ul{178.36}_{(34.16)}$ & $\ul{83.92}_{(3.02)}$ & $\ul{207.11}_{(36.51)}$ & $\ul{204.41}_{(4.63)}$ & ${301.61}_{(65.76)}$ & $\ul{254.11}_{(4.92)}$ & ${334.06}_{(74.09)}$\\
        CASTANET:NoARL2 & ${48.41}_{(1.44)}$ & ${192.79}_{(28.69)}$ & ${94.04}_{(4.87)}$ & ${223.15}_{(28.13)}$ & ${237.99}_{(26.32)}$ & ${350.85}_{(49.88)}$ & ${294.45}_{(32.05)}$ & ${389.07}_{(51.94)}$\\
        CASTANET(Proposed Model) & $\bm{41.83}_{(2.41)}$ & $\bm{160.34}_{(26.09)}$ & $\bm{80.94}_{(2.87)}$ & $\bm{186.40}_{(28.04)}$ & $\bm{200.98}_{(3.99)}$ & $\bm{249.09}_{(46.93)}$ & $\bm{250.69}_{(4.35)}$ & $\bm{270.05}_{(51.13)}$\\
    \bottomrule
    \end{tabular}}
\end{table*}
}

\subsection{Metrics}
We evaluate the congestion prediction performance of the model using RMSE:
\begin{align} \label{eq:metric}
    &{\rm RMSE} = \sqrt{\frac{1}{\tau T' N}\sum_{t=1}^{\tau}\sum_{i=1}^{N}\sum_{t'=1}^{T'}\bigl(\hat{Y}_{i,t+t'}-Y_{i,t+t'}\bigr)^2}. 
\end{align}
For a test dataset $D=\{(\bmX_t, \bm{A}_{t-1}, \bm{a}_t, \bm{Y}_t, \bm{W})\}_{t=1}^{\tau}$ consisting of observations for $\tau$ time steps, we measure the prediction
performance after an incident, such as accidents. 
Let $S=\{(i,t)\mid 1\le i\le N,\,1\le t\le \tau\}$.
We define the incident subset as
$D'=\{(i,t)\in S\mid a_{i,t}=1\}$, and report RMSE on $D'$ as RMSE (Incident). 

Since the queue length data are zero-inflated, a model that predicts zero for all data points may \KT{achieve artificially high performance}. 
To address this, we divide the test dataset into four subsets and evaluate the performance for each:
{
\setlength{\leftmargini}{15pt}
\begin{itemize}
    \item \textit{All:} All data points.
    \item \textit{Non-zero:} Data points where queue length is not zero.
    \item \textit{Top 10\%:} The top 10\% of data points in terms of queue length among the non-zero data points.
    \item \textit{Top 5\%:} The top 5\% of data points in terms of queue length among the non-zero data points.
\end{itemize}
}

\subsection{Baseline models}
We employ four conventional models as baselines, including historical averages (HA), Lasso, Ridge, and XGBoost.
We additionally use seven state-of-the-art traffic prediction models: DCRNN~\cite{li2018dcrnn}, GWNet~\cite{wu2019gwnt}, AGCRN~\cite{bai2020agcrn}, MegaCRN~\cite{jiang2023megacrn}, QTNet~\cite{shirakami2023qtnet}, PDG2Seq~\cite{FAN2025106941} and STAEFormer~\cite{liu2023staeformer}. 
All information is concatenated along the channel direction and fed into each model. 
The output is the queue length for the next hour. 
Based on~\cite{shirakami2023qtnet}, for models other than QTNet, the loss function uses only the terms related to queue length from Eq. \eqref{rq:loss_detail_1} and Eq. \eqref{rq:loss_detail_2} , excluding the weight coefficient $r_{i,k}$.
We also evaluate two variants of CASTANET without adversarial representation learning ($\lambda_{\rm Conf}=0$):
(i) NoARL1, which applies inverse-propensity weighting to Eq.~\eqref{eq:loss_y}  to mitigate treatment-selection bias, and
(ii) NoARL2, which reweights Eq.~\eqref{eq:loss_y} by the inverse empirical incident frequency to alleviate incident sparsity.

\subsection{Experimental Settings}\label{sec:setting}
Hyperparameters are tuned to improve the accuracy of queue length on the validation dataset. We set $d_{\rm emb}=16$, $d_{\rm rep}=40$, and $d_{\rm node}=10$. 
The parameters for the loss function were set to $\lambda_1=4.0$, $\lambda_2=0.02$, $\lambda_3=0.05$, $\lambda_4=0.08$, $\delta_{1}=100$, $\delta_{2}=100$,
$\lambda_{\rm Conf}=0.1$, $\alpha_1=10$, and $\alpha_2=0.01$. The loss weights were set so that the magnitudes of the queue-length, normalized queue-length, speed, and flow loss terms are comparable on the validation set, preventing any single term from dominating optimization. Additionally, curriculum learning was applied to efficiently train the outcome predictor network using the QT-layer, \KT{following} the setup described in~\cite{shirakami2023qtnet}. For model optimization, we used Adam, with the learning rate starting at 0.001 and decaying by a factor of 0.97 each epoch. Training was conducted for 200 epochs, with early stopping applied if the validation loss did not improve for 10 consecutive epochs. All experiments were conducted using Python 3.8.10 and PyTorch 1.13.1 on a server with an NVIDIA A100 GPU.
We needed propensity scores $\bm s_{i,t}$ for CASTANET:NoARL1. We built a neural network to predict propensity scores given $\bmX_t$ and $\bm{A}_{t-1}$. 
Trained to minimize Equation \eqref{eq:loss_a}, this model achieved an AUC of 0.84 for current incident status prediction, indicating the validity of the estimated propensity scores.

\begin{figure}[t]
    \centering
    \includegraphics[width=\linewidth]{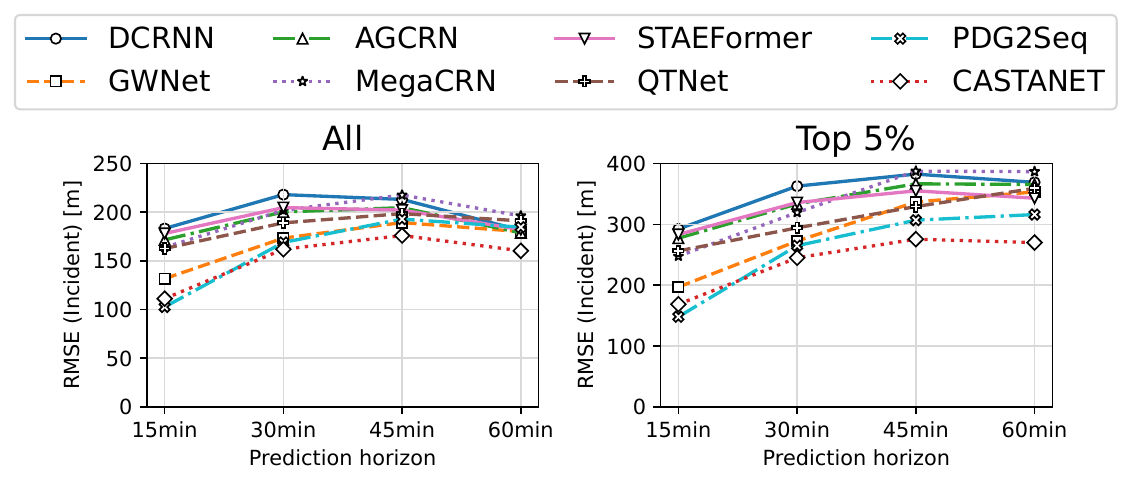}
    \caption{RMSE (Incident) for one-hour-ahead predictions.}
    \label{fig:rmse_acc_baselines}
\end{figure}

\subsection{Results}
To address \textbf{RQ1}, we evaluated the average and standard deviation of RMSE across the datasets (a)--(e).
We show the results in Table ~\ref{tab:rmse_basleine}. 
We confirmed that CASTANET improved queue length predictions regardless of whether an incident occurred.
In the All setting, CASTANET reduces RMSE by 4.0\% compared with the best baseline (QTNet),
and reduces RMSE (Incident) by 10.1\% compared with the best competing method (CASTANET:NoARL1).
Under severe congestion (Top 5\%), CASTANET outperformed the state-of-the-art models by a wide margin of 14.55\% in RMSE (Incident), which demonstrates the effectiveness of our proposed architecture for utilizing incident data. CASTANET also outperformed NoARL1 and NoARL2, indicating that ARL provides additional benefits beyond inverse-propensity weighting and incident-frequency reweighting in our setting. 
NoARL2 degrades performance compared to CASTANET because aggressive weighting of the
rare treated class increases variance. Unlike sample reweighting, ARL explicitly enforces treatment-invariant representations, which helps the model utilize incident information. HA achieves comparable RMSE to STGNN-based baselines, but performs poorly on RMSE (Incident) particularly in Top 10\% and Top 5\%. This result indicates that most traffic patterns are periodic. 
However, sudden incidents cause non-periodic congestion, which is difficult for HA to predict. Despite this difficulty, CASTANET can predict both periodic and non-periodic congestion. PDG2Seq utilizes dynamic graph convolution, which is particularly effective when accidents occur, resulting in the second-best performance.
Figure \ref{fig:rmse_acc_baselines} shows the progression of RMSE (Incident) up to one-hour-ahead for CASTANET and baseline STGNN-based models. 
The performance gain of CASTANET persists over longer horizons, suggesting that it captures not only the immediate post-incident impact but also the sustained congestion dynamics post-incident. PDG2Seq outperforms CASTANET in the 15-min-ahead prediction setting because
PDG2Seq utilizes dynamic graph convolution.

\begin{figure}[t]
    \centering
    \includegraphics[width=1.0\linewidth]{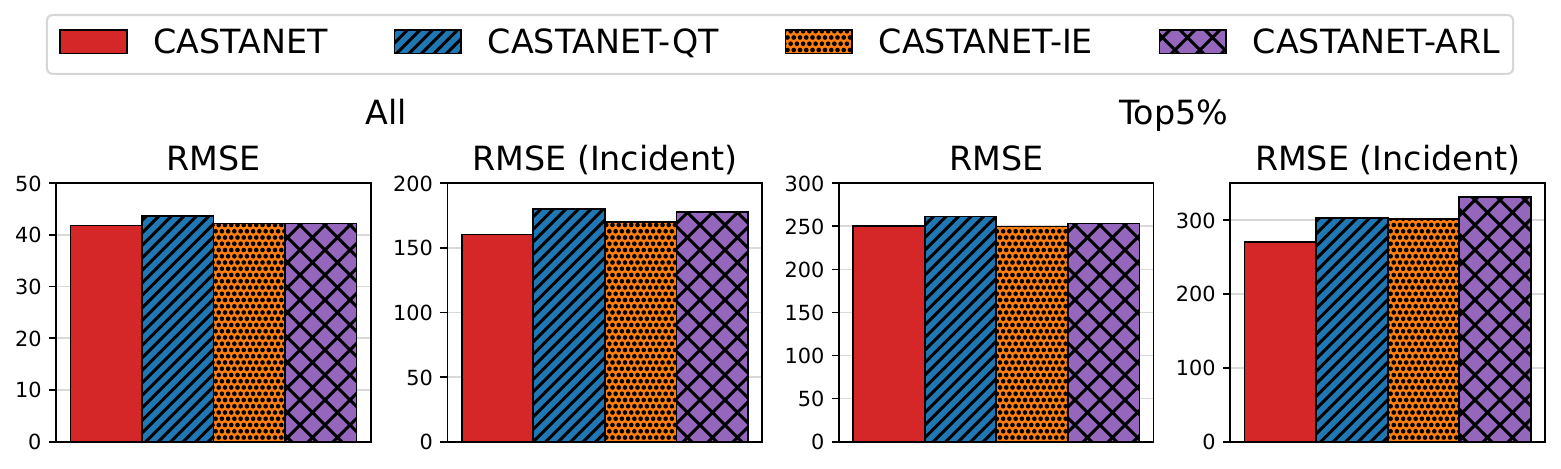}
    \caption{
    Ablation study on RMSE and RMSE (Incident).
    }
    \label{fig:ablation_rmse}
\end{figure}

\subsection{Ablation Study}
To address \textbf{RQ2}, we conducted an ablation study comparing CASTANET with its variants; \textbf{CASTANET-IE}: CASTANET without the incident embedding, \textbf{CASTANET-QT}: CASTANET without the QT-layer, and \textbf{CASTANET-ARL}: CASTANET without ARL.
Figure \ref{fig:ablation_rmse} \KT{presents} the one-hour-ahead RMSE and RMSE (Incident) averaged over the datasets (a)--(e) for CASTANET and its variants. Compared to CASTANET, all variants yielded a performance decline, indicating the contribution of each component to improving prediction accuracy.
Of these components, ARL contributed the most to the improvement in performance, particularly with regard to RMSE (Incident).
Because ARL mitigates bias and overfitting in the incident distribution of the training data, we expect it to generalize better by providing unbiased predictions on the test data.
The QT-layer also enhances prediction accuracy due to its traffic engineering constraints.
Incident embedding consistently improves performance across all subsets, indicating that it effectively conveys incident information to the model.

\subsection{Qualitative Evaluation}
\begin{figure}[t]
    \centering
    \includegraphics[width=0.9\linewidth]{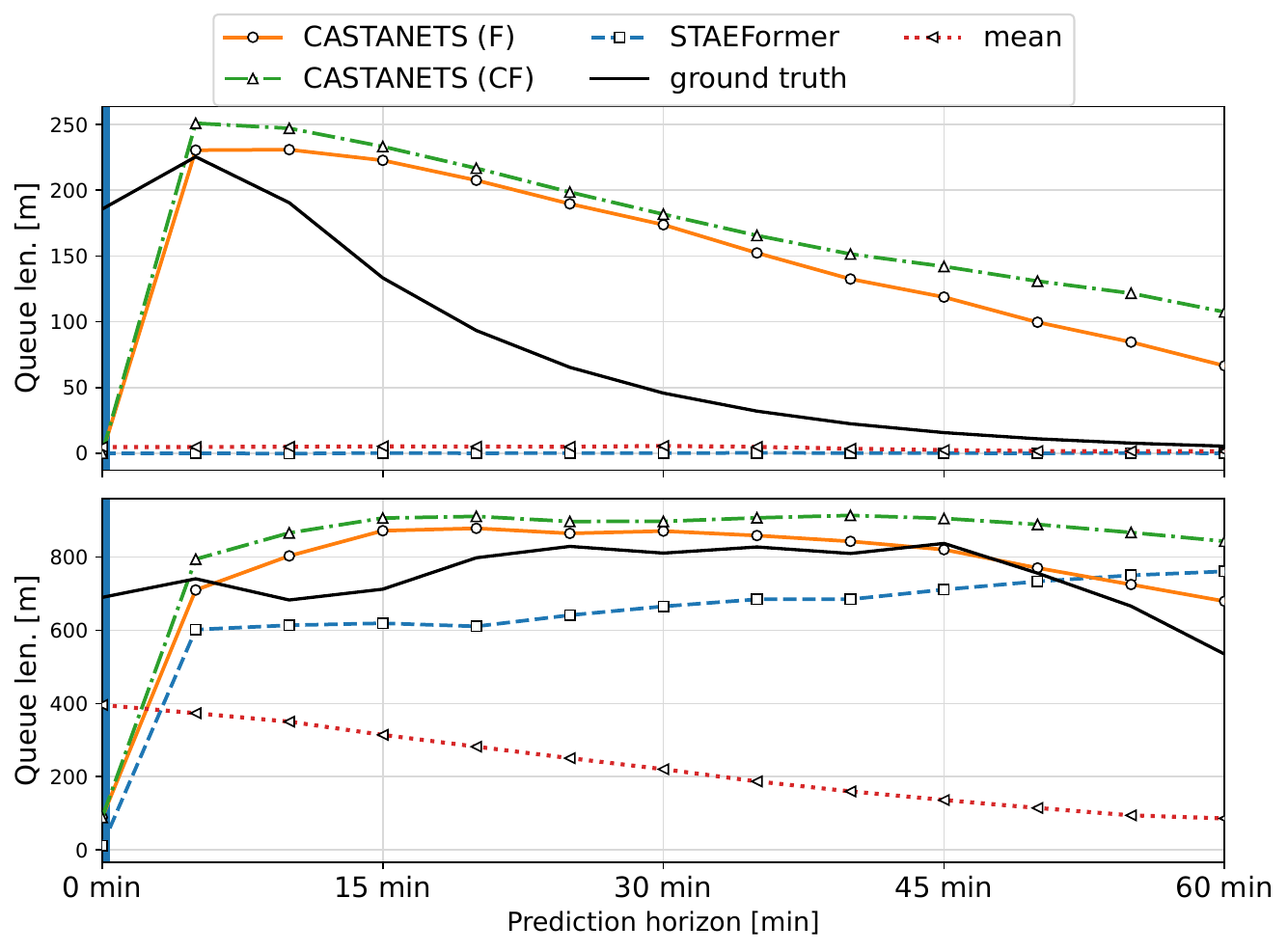}
    \caption{Every five-minute factual and counterfactual predictions by CASTANET. The incident time is indicated by the blue vertical line at 0 min on the x-axis.}
    \label{fig:cf}
\end{figure}

To address {\bf RQ3}, we visualize the predicted outputs of CASTANET.
Since CASTANET is a model based on causal inference methods, we visualize the factual prediction (CASTANET (F)) where an incident is provided as an input, and the counterfactual prediction (CASTANET (CF)), in which we set the current incident indicator to zero. We show every five-minute predictions up to one-hour-ahead in Figure~\ref{fig:cf}.
For simplicity, STAEFormer is shown here for a comparison. In both cases, CASTANET (F), which uses factual incident data, provides more accurate predictions than CASTANET (CF), which uses counterfactual incident data. Moreover, the effect of the incident indicator varies across cases, suggesting that CASTANET adjusts its response to incidents according to the traffic context rather than applying a fixed rule.

\subsection{Computation Time}

To address {\bf RQ4}, we measure the computational time of CASTANET and STAEFormer on a single NVIDIA A100 GPU. For the Tokyo dataset, CASTANET requires 0.02 seconds to generate multi-horizon predictions from 5 to 60 minutes ahead and training for 84 epochs took 641 minutes. STAEFormer requires 0.05 seconds to generate predictions and training for 20 epochs took 656 minutes. This inference and training time are comparable to that of the STGNN-based baseline.  In our current setting, CASTANET uses approximately 300 MB of GPU memory. We believe that CASTANET can be applied to a significantly larger road network or city-wide traffic control system.

\section{Conclusion}
In this paper, we proposed CASTANET, which integrates an STGNN backbone with balanced representation learning from the treatment effect estimation framework to predict traffic congestion under sudden incidents.
In future work, we plan to incorporate incidents beyond accidents,  to examine the sensitivity to violations of the underlying causal assumptions and to introduce an off-policy evaluation metric.

\section*{ACKNOWLEDGMENTS}
The authors would like to thank the Traffic Facilities and Control Division,
Traffic Bureau, Tokyo Metropolitan Police Department for providing the traffic data.
This work was supported by JST FOREST (JPMJFR232S), JST BOOST (JPMJBY25D0),
and JSPS KAKENHI (26K02984).

\bibliographystyle{IEEEtran}
\bibliography{reference}

\end{document}